# Teaching large language models to reason like expert diagnosticians

Thomas A. Buckley[1], Riccardo Conci[1], Peter G. Brodeur[2], Jason Gusdorf[2], Sourik Beltrán[2,3], Bita Behrouzi[2], Byron Crowe[2], Jacob Dockterman[4], Muzzammil Muhammad[2], Sarah Ohnigian[2], Andrew Sanchez[2], James A. Diao[1,5], Aashna P. Shah[1], Daniel Restrepo[6], Eric S. Rosenberg[7], Andrew S. Lea[8], Emily Glanton[1], Kimberly LeBlanc[1], Undiagnosed Diseases Network, Marinka Zitnik[1,9], Scott H. Podolsky[10,11], Zahir Kanjee[2], Raja-Elie E. Abdulnour[12], Jacob M. Koshy[2], Adam Rodman[2], and Arjun K. Manrai[1]*

1. Department of Biomedical Informatics, Harvard Medical School, Boston, MA
2. Department of Medicine, Beth Israel Deaconess Medical Center, Boston, MA
3. The Mongan Institute, Massachusetts General Hospital, Boston, MA
4. Division of Gastroenterology, Brigham and Women's Hospital, Boston, MA
5. Department of Medicine, Brigham and Women's Hospital, Boston, MA
6. Department of Medicine, Massachusetts General Hospital, Boston, MA
7. Department of Pathology, Massachusetts General Hospital, Boston, MA
8. Department of Health Humanities and Bioethics, University of Rochester School of Medicine and Dentistry, Rochester, NY
9. Kempner Institute for the Study of Natural and Artificial Intelligence, Harvard University, Cambridge, MA
10. Center for the History of Medicine, Countway Library of Medicine, Harvard Medical School, Boston, MA
11. Department of Global Health and Social Medicine, Harvard Medical School, Boston, MA
12. Division of Pulmonary and Critical Care Medicine, Brigham and Women's Hospital, Boston, MA

* Correspondence: Arjun_Manrai@hms.harvard.edu

## ABSTRACT

Differential diagnosis is an iterative process that integrates patient information with broader medical knowledge[1]. Clinical case series such as the *NEJM* Clinicopathologic Conferences (CPCs), published continuously since 1923[2], feature expert physicians who demonstrate diagnostic reasoning to peers, and have been used for decades to evaluate AI. However, prior AI evaluations have largely focused on final diagnostic accuracy rather than nuanced clinical reasoning[3,4]. Here, we introduce Dr. CaBot, an agentic AI system that emulates an expert diagnostician by generating written and narrated slide-based presentations from an initial case description alone. CaBot recently generated the first AI diagnosis published in the 100+ year history of the *NEJM* CPCs[5]. In blinded evaluations, physicians misclassified the source of the differential (CaBot vs. physician-written) in 46/62 (74%) of trials and rated them favorably across quality dimensions. When tasked with solving cases for 72 patients with undiagnosed disease from the NIH Undiagnosed Diseases Network[6], CaBot identified the working diagnosis in 50/72 (69%) of cases from referral notes alone. To promote transparency and research, we also developed CPC-Bench, a physician-validated benchmark based on 7,102 CPCs and 47,648 questions across 10 tasks. We show that CaBot outperforms frontier models on CPC-Bench, and release both CaBot and CPC-Bench publicly to foster progress in clinical AI.

## MAIN

For the past half century, artificial intelligence (AI) tools have been evaluated using clinical case series[7]. In particular, solving the *NEJM* Clinicopathologic Conferences (CPCs), introduced in 1923 by Dr. Richard Cabot and held at Massachusetts General Hospital, has long been seen as the gold standard for assessing the quality of differential diagnosis generation by AI systems[3,4,8–13]. However, existing evaluations using these cases have considerable limitations because the CPCs are pedagogical in nature; despite being real cases, they are highly curated and information dense, intended to demonstrate an idealized process of expert reasoning rather than real-life diagnosis. Moreover, almost all evaluations have been limited to focusing on the final diagnosis[3,4,11]. This does not measure the actual work of the expert physician in the case, who not only provides a final diagnosis but more importantly argues a nuanced and complete differential diagnosis grounded in the clinical literature, and demonstrates their thinking to a live audience of expert colleagues. The value of the CPC is the process—the ability to integrate new information, reference guidelines and research studies, process conflicting diagnostic data, and to explicitly discuss these diagnostic strategies, rather than merely deliver a final diagnosis.

To extend the abilities of clinical AI beyond final diagnosis generation, here we introduce “Dr. CaBot”, an agentic AI system designed to emulate the reasoning and presentation of an expert physician diagnostician. Dr. CaBot is designed to not only provide a final diagnosis but also articulate intermediate steps in both written and live presentation form. Given only the initial case presentation and imaging studies, the system constructs a written discussion with citations to the clinical literature, proposes diagnostic and management strategies, and prepares and narrates a slide-based video presentation to justify its reasoning. We evaluated Dr. CaBot against thousands of original expert discussants that presented cases published over the past century in the *NEJM*, and on 72 cases submitted to the Undiagnosed Diseases Network (UDN), a multicenter National

Institutes of Health (NIH) research study that supports patients with rare and previously undiagnosed conditions.

To further promote community benchmarking beyond final diagnostic accuracy of large language models (LLMs), we introduce CPC-Bench, a large-scale, publicly available, and physician-validated benchmark based on the CPCs. CPC-Bench is based on 7000+ CPCs and includes 47,648 questions spanning 10 benchmark tasks such as differential diagnosis generation, generation of testing and management plans, literature search, information omission, and visual differential diagnoses. We evaluate CaBot and several frontier large language models against CPC-Bench. We publicly release both Dr. CaBot and CPC-Bench online at cpcbench.com, aiming to foster standardized and nuanced evaluations of AI, and to catalyze AI development.

# RESULTS

## CaBot: An AI Expert Discussant

We created Dr. CaBot, a multimodal and interactive AI expert discussant that can respond to a case presentation by generating a text differential diagnosis along with a video containing a narrated presentation of the differential diagnosis (Fig. 1). Dr. CaBot is an agentic system that equips a reasoning model (e.g., OpenAI o3 or GPT-5.4) with external tools. When provided with a case, Dr. CaBot retrieves two similar cases from a set of over 6,000 clinical cases as examples of expert clinical reasoning. The model then iteratively searches a set of 1,623,047 clinical abstracts to develop a differential diagnosis with citations to the clinical literature. CaBot makes up to 25 search queries to the database, with 5 articles returned each time, ranked by semantic similarity (Fig. 1B, Fig. 1C). This text is then used to generate a complete slideshow presentation, by using an LLM (OpenAI o3) to generate LaTeX Beamer code paired with a slide-by-slide narration (Fig. 1D). CaBot can rapidly respond to follow-up questions with tool use disabled (Fast mode, Fig. 1). We generated sample video and text case conferences for 15 clinical cases, available at https://cpcbench.com/discussant.html.

### Blinded Physician Evaluations of CaBot Performance

We provided five internal medicine physicians with paired differentials in randomized order (one CaBot, one *NEJM* discussant) from 31 cases, yielding 62 differential diagnoses. Physicians were blinded to the source (human vs. AI) of the differential. Two example discussions from the physician discussant and CaBot are shown in Fig. 2A. The physicians correctly identified the source of the differential in 26% (16/62) of trials (95% CI: 17% to 38%) (Fig. 2B). One case was graded as unsure and counted as misclassified. We collected qualitative rationales for why physicians thought a differential was AI or human, which showed frequent confusion. For one case, the physician commented that CaBot's text exhibited "Human speech patterns, well reasoned, very relevant to history" while the physician-generated text had "AI speech patterns."

Next, while still blinded to the source of the differential, physicians scored each text for a subset of 44 trials based on 22 cases on a Likert scale assessing the overall quality of the differential diagnosis, justification of each item on the differential, quality and accuracy of the citations, and learner engagement. Across 44 trials and all four metrics, physicians rated Dr. CaBot as consistently better than the original *NEJM* discussant (Fig. 2C).

### CaBot on Unsolved Rare Disease Cases

We evaluated CaBot (with GPT-5.4 as base model) on 72 de-identified clinical summary letters written by health care providers for their patients at the time they were referred to the Undiagnosed Diseases Network (UDN), an NIH clinical research study for patients with rare conditions who remain undiagnosed despite extensive prior evaluation. Of these, 26 eventually reached a diagnosis through UDN evaluation and 46 had a suggested diagnosis based on case review, which involved analysis of all available records of the referred patient.

Across the 72 cases, CaBot's diagnosis named the correct broad diagnosis category in 50 cases (69.4%), mentioned the specific diagnosis or causal gene in 46 cases (63.9%), and selected the correct diagnosis as the top pick in 35 cases (48.6%). Among the 26 cases with a diagnosis identified via UDN evaluation, using the referral letter alone, CaBot identified the correct diagnosis category in the differential in 16 (61.5%) cases, and as the top pick in 14 (53.8%) cases. CaBot named the specific diagnosis or gene on the differential in 13 (50.0%) cases, and as the top pick in 12 (46.2%) cases. Among the 46 cases that had a suggested diagnosis based on case review alone, CaBot's diagnosis was concordant with the UDN site's leading hypothesis in 34 cases (73.9%), it named the specific diagnosis or gene hypothesized by the UDN in 33 cases (71.7%), and it placed it as the top pick in 21 cases (45.7%) (Table 1).

In another subset of 26 cases without mention of the diagnosis in the referral letter, CaBot identified the correct broad diagnosis category in 10 (38.5%), named the specific diagnosis or causal gene in eight (30.8%), and selected the correct diagnosis as its top pick in six (23.1%).

**CPC-Bench: A New Benchmark for Expert Clinical Reasoning**

To promote transparency and evaluate CaBot and other LLMs at scale, we introduce CPC-Bench, a new benchmark of clinical reasoning based on 7,102 CPCs which includes 47,648 questions and 10 tasks spanning diagnosis, management, clinical literature retrieval and medical image analysis (Fig. 3). The complete list of tasks and evaluation metrics is provided in Table 2. CPC-Bench was developed using thousands of clinical cases and images from *NEJM* Clinicopathologic Conferences, which are complex case presentations seen at Massachusetts General Hospital with a physician differential diagnosis and subsequent documented management steps, as well as the complete *NEJM* Image Challenges. 10 physicians annotated the CPCs to develop a benchmark for LLMs at key moments in the patient course or when key information is missing (ex. normal findings explicitly asked for by the physician).

**Comparing CaBot and Frontier Models on CPC-Bench**

We measured the overall performance of CaBot compared to frontier LLMs across tasks from CPC-Bench. On benchmark tasks requiring only text, such as diagnosis using the initial case presentation, or finding the correct citation for a clinical claim, CaBot answered 81% of questions correctly (95% CI, 79% to 82%), outperforming Gemini 2.5 Pro (76%, 95% CI, 75% to 78%), o3 (76%, 95% CI, 75% to 78%), and Claude-4-Sonnet (71%, 95% CI, 69% to 72%) (Fig. 4A). On all multimodal tasks, such as the *NEJM* Image Challenge and image-only CPCs, CaBot achieved 68% accuracy (95% CI, 66% to 69%), outperforming o3 (67% [95% CI, 65% to 69%]) and Gemini 2.5 Pro (67% [95% CI, 65% to 69%]) (Fig. 4B).

We then analyzed the diagnostic performance of two general-purpose LLMs, o3 and GPT-4o, provided key sequential touchpoints within the initial case presentation, obtained by a physician-validated annotation pipeline (Fig. 4C). Across the 377 cases, there was an average of 8.1 events per case (SD= 3.9). Both o3 and GPT-4o showed stepwise gains in diagnostic accuracy at each event. From Event 1 to Event 5, accuracy increased by 28 percentage points for o3 (95% CI, 20% to 35%) and 15 percentage points for GPT-4o (95% CI, 9% to 22%). After the first event, the correct pathological diagnosis appeared somewhere in the differential in 27% of cases for o3 (95% CI, 23% to 32%) and 24% for GPT-4o (95% CI, 20% to 29%). The mean rank of the correct diagnosis improved from the first to the final event by 0.30 positions for o3 and 0.48 positions for GPT-4o (lower rank indicates better placement); at every event, the mean rank remained lower for o3 than for GPT-4o.

Through the case annotations, each key piece of investigative information in the case was categorized as ‘abnormal’ or ‘explicitly not abnormal’. This latter definition is key to differentiate information that a clinician has to explicitly look for and ask and that the patient would not directly offer, such as ‘there was no lymphadenopathy’, or ‘the CT scan did not show […]’. When omitting these key normal findings we note that the performance for top-1 diagnosis drops on average around 2-3% across models (Supplementary Table 1).

**Comparing CaBot and Frontier Models on Clinical Literature Retrieval**

In the Literature Search task, we provide models with a snippet of text containing a clinical claim (ex. “A gastrin level that increases by more than 120 pg per millilitre in response to secretin is diagnostic of a gastrinoma [CITATION_HERE]”), as well as the date this claim was published, using 659 clinical claims published after 2000. We then measured accuracy by checking whether the correct citation appeared in the top 10 suggestions from the model. We configured CaBot with up to 5 uses of the clinical database we developed, which contains papers from all journals included in the retrieval benchmark. As shown in Figure 4D, CaBot retrieved the correct primary source in its top 10 responses for 54% of questions (95% CI, 50% to 58%), compared to 27% (95% CI, 24% to 31%) for the base o3 model without retrieval. Gemini 2.5 Pro retrieved the same primary source in its top 10 responses for 44% of questions (95% CI, 40% to 48%), exceeding the performance of o3 and Claude 4 Sonnet. Other models performed poorly (10% accuracy or less). Citation accuracy declined for articles published before 1990 and for those published after 2020 (Supplementary Fig. 1).

**Image Interpretation by Specialty**

We assessed the capabilities of CaBot and other LLMs for medical imaging interpretation using a subset of *NEJM* Image Challenge cases that include only an image and ask “What is the diagnosis?” without other textual information (n = 230). Additionally, we used questions from the Visual Question-Answering (VQA) task, which is constructed from image–caption pairs extracted from the CPCs, and measures the ability of a model to perform direct image interpretation (ex. “Which morphological change is shown in Panel A?”). Across benchmarks, we measured performance on tasks in dermatology, radiology, cardiology, and histology. Models performed best on dermatology (CaBot accuracy 79%, 95% CI, 69% to 87%) and histology (CaBot accuracy 76%, 95% CI, 73% to 79%), with worsened performance in radiology (CaBot accuracy 58%–67%) and cardiology (CaBot accuracy 58%, 95% CI, 47% to 68%). CaBot exceeds or performs on par with o3 and Gemini 2.5 Pro across clinical specialties (Fig. 4E).

**Using Dr. CaBot and CPC-Bench**

We have released Dr. CaBot publicly at https://cpcbench.com/discussant.html. The site is preloaded with video and text case conferences for 15 cases published between 2000 and 2025. Users can also generate new video and text case discussions. Additionally, we have created a public leaderboard of model performance across all 10 tasks in CPC-Bench, available at https://cpcbench.com. We have also made available a submission portal for new models to be evaluated on CPC-Bench. By registering at https://cpcbench.com/submit.html, researchers can submit custom models for evaluation on the private CPC-Bench benchmark. Furthermore, upon publication, we will release a publicly-available dataset of 100 modern cases, used with permission from *NEJM*, which includes 865 clinical events and 4,735 derived benchmark questions.

## DISCUSSION

Most studies of the diagnostic reasoning abilities of LLMs and other AI models have assessed performance primarily through evaluating final diagnostic accuracy on clinical vignettes, or multiple-choice questions on standardized exams. This has rightly attracted criticism. The actual diagnostic process entails communication, integrating conflicting information, and navigating uncertainty[14,15].

To advance the evaluation of the reasoning capabilities of LLMs, we developed Dr. CaBot, an agentic AI system that emulates an expert physician in written and live presentation form. Beyond achieving *NEJM* discussant-level final diagnostic accuracy on complex cases, CaBot generates coherent clinical reasoning of the kind expected in clinical practice. In blinded evaluations, physicians were unable to distinguish CaBot-generated discussions from the original *NEJM* case discussions, indicating that the system can emulate the style and structure of expert clinical reasoning, not only its conclusions. We further evaluated CaBot on challenging rare disease cases referred to the NIH Undiagnosed Diseases Network (UDN), where it identified the correct diagnosis in the majority of cases from the referral letter alone. Taken together, these results suggest that LLM-based systems may provide rapid, useful second opinions for clinicians to consider for patients with undiagnosed conditions, potentially helping shorten the years-long delays these patients typically experience today.

CaBot builds upon recent progress in developing agentic systems for diagnosis of complex cases, including DeepRare[13] and RareAgents[16]. Whereas prior systems have been designed and evaluated in their ability to generate a top-K ranked list of diagnoses, CaBot was designed in close collaboration with practicing physicians to emulate the actual reasoning process of an expert diagnostician. CaBot builds a coherent clinical argument, weighs competing hypotheses, and grounds each step in the high-impact clinical literature. Rather than ranking entries from a disease ontology, CaBot produces the kind of reasoned discussion an expert would deliver to peers, both as written text and as a narrated, slide-based video case conference. This design was iteratively refined on held-out cases based on direct physician feedback, and the resulting reasoning was shown to surpass on several quality dimensions that of physicians selected to present at the Massachusetts General Hospital.

To conduct nuanced evaluations of clinical text-based and multimodal capabilities at scale, we introduced CPC-Bench, a physician-validated benchmark to standardize evaluation of AI models against the many types of expert reasoning contained in these cases. When applying current AI models to CPC-Bench, we found widespread performance differences across literature retrieval, image interpretation, and differential diagnosis generation with varying information. We are releasing both CPC-Bench and CaBot publicly as community resources to foster innovation and track progress in AI.

Our study has implications for the future of human-AI collaboration in diagnostic reasoning. Experimental studies suggest minimal benefit from physician use of LLMs when used for purely diagnostic classification tasks where they receive a final diagnosis only[17] – that is, where physicians solely receive diagnostic suggestions. However, psychological studies of human-human collaboration in diagnosis suggest that when reasoning can be exposed, not just final diagnosis, it may provide a pathway to improving diagnostic performance[18–20]. Our study showcases new avenues of allowing clinicians to interact with AI in the diagnostic process, potentially serving to increase performance and build physician trust. This more accurately reflects the nature of physician-based second opinions ("curbside consults")[21]. Future studies will need to more explicitly study the impacts of human-AI collaboration and how and when to trigger AI second opinions in actual clinical care.

Our study has several limitations. First, the training data used for the LLMs evaluated here are not publicly known and may include some of the CPCs along with their answers. We analyzed the performance on newer cases which are unlikely to be accessed by the model and found similar performance, reducing concerns of leakage (Supplementary Table 2). Second, CPCs are highly curated, with a bias towards rare diseases or unusual presentations of common diseases. However, this may make applications such as the UDN well-suited for such systems. Third, the 10 tasks evaluated in CPC-Bench are not a comprehensive set of tasks in clinical medicine. None directly measured long-context reasoning or clinical summarization, which is another major challenge for clinical AI models,[22] nor do they assess structured data commonly found in the EHR[23]. Fourth, several tasks in CPC-Bench do not have a large human baseline against which to compare AI performance[11]. Finally, most of our annotations and comparisons were done by internal medicine physicians, and future involvement of other disciplines and sub-specialists would help assess generalizability.

Large language models now rival physician experts in text-based diagnostic reasoning and can generate presentations often indistinguishable from human work. These abilities extend to real cases of rare disease patients, potentially helping shorten diagnostic odysseys these patients often experience. At the same time, LLMs still struggle with image interpretation and multimodal integration. By releasing Dr. CaBot, CPC-Bench, and public leaderboards, we offer a path to analyze clinical reasoning skills of LLMs and to track progress on clinical reasoning across decades of cases. In doing so, we extend Dr. Richard Cabot's vision of the CPC as both a teaching tool and a benchmark for clinical reasoning—now for humans and machines alike.

# METHODS

## Dr. CaBot

*Clinical Literature Search*

We cloned the entire OpenAlex dataset[24], an index of published literature sourced from PubMed, Crossref, and other sources. We subset the dataset to include any article published in any clinical journal with an impact factor of at least 10, resulting in 204 total journals (using Clarivate 2024 JIF, see Supplementary Information for complete list). This dataset includes 3,474,245 total papers, 1,623,047 of which contain an abstract (46.7%). Using the OpenAI text-embedding-3-small model, we embedded all abstracts in their entirety, truncating abstracts longer than 8000 characters and excluding abstracts shorter than 100 characters. These were stored alongside other features in PostgreSQL using an Inverted File Flat Index (IVF Flat) with pgvector – an approximate nearest neighbors algorithm. For querying, we rank candidate abstracts by their Euclidean distance to the query embedding as returned by the IVF Flat index. Since some case presentations from *NEJM* appear in this database, we exclude abstracts by case DOI.

*CaBot System*

CaBot is an agentic system based on a general-purpose reasoning model that answers clinical questions and generates differential diagnoses in the style of a CPC discussant, in both written and video presentation form. In this study, we use OpenAI o3 as the base model in all experiments, except the rare disease evaluation, in which we use GPT-5.4. In the initial prompt to the agent, we provide two similar "Differential Diagnosis" sections to mimic the style of the expert discussant (described below). In subsequent iterations, the system iteratively queries the clinical literature tool (described above). The model can make up to 25 queries to the clinical literature tool.

To mimic the style of the expert discussant, CaBot identifies the two CPC "Presentation of Case" sections most similar to the index case by searching embeddings of over 6,000 cases available through May 31, 2024 (the o3 pretraining cutoff), excluding the index case itself. By default, the comparators are modern cases (2000 onward) published within two years of the index case. The o3 model then uses only the "Differential Diagnosis" sections from the two selected cases, along with instructions to reproduce their style. CaBot was iteratively refined on 9 held-out cases based on physician feedback to more closely emulate an expert discussant. Prompts were crafted to discourage the use of bullet points and to omit concluding or introductory sections (e.g., "Synthesis," "Conclusion"), thereby matching the format of published CPC differentials. Complete system prompts are provided in the Supplementary Information.

*Generating Video Case Conferences with Dr. CaBot*

Given the case text and images, we prompt o3 to generate LaTeX Beamer slides (PDF) with a slide-by-slide script. Narration was synthesized using an OpenAI text-to-speech model (TTS-1 HD), and the slides and audio were combined into a video with FFmpeg. We iteratively tuned the prompt to be more human-like in narration, including by instructing the model to explain itself ("signpost") as expert discussants often do, and to use thinking words like "uh" and "um."[25] Prompts are available in the Supplementary Information.

**Blinded Physician Evaluation of CaBot**

For the physician evaluation, we first selected all CPCs published between 2023 and February 27, 2025. We excluded cases used for model development (9 cases). Additionally, we excluded cases from evaluation if they were management-focused, or if the physician discussant knew the final diagnosis, participated in the care of the patient, presented figures or tables during their differential diagnosis, or used information not available in the presentation of case section (such as additional testing or treatment details). From the remaining set, we selected the 29 most recent CPCs to reduce the risk of model memorization.

Five internal medicine physicians (two attendings, three residents) reviewed paired differentials from Dr. CaBot and the original *NEJM* discussant, blinded to the source, for 31 cases yielding a total of 62 trials. Out of the 29 available, responses to 27 unique CPCs were scored, with 4 CPCs scored by two physicians. We did not receive responses for the other 2 cases. Reviewers were asked to identify which written differential diagnosis was AI- vs. human-generated, with the order of differentials randomized. For a subset of 22 cases (44 trials), physicians also rated both CaBot and human differentials using a rubric on 1-5 Likert scales for: (1) Overall differential quality, (2) Diagnostic justification, (3) Citation quality, and (4) Learner engagement.

**CaBot for Rare Disease**

We retrieved 84 de-identified clinical summary letters prepared by health care providers for patients being considered by the Undiagnosed Diseases Network (UDN), a multicenter National Institutes of Health (NIH) clinical research study that seeks to identify diagnoses for patients with rare and previously undiagnosed conditions. Each UDN site accepts applications from patients with complex medical presentations who have remained without a unifying diagnosis despite extensive prior evaluation. Twelve cases were flagged for exclusion by certified genetic counselors (K.L., E.G.) because the materials were too short to contain meaningful clinical information or because a ground-truth diagnosis was not available, yielding 72 cases for analysis (26 with a diagnosis made through UDN clinical and/or research evaluation; 46 that had a diagnosis suggested based on medical record review only).

Each case consisted of a de-identified clinical summary letter. Blinded to the final diagnosis or UDN analysis, we provided these materials to Dr. CaBot and requested a comprehensive differential diagnosis. All authors were blinded to the final diagnosis during CaBot trials, except the certified genetic counselors (K.L., E.G.). One genetic counselor assessed each CaBot letter against the final UDN diagnosis where available, or against the UDN site's leading candidate genes or disease hypotheses otherwise. As needed, cases were discussed between the genetic counselors to reach consensus during this assessment (e.g., cases where a patient received more than one diagnosis, suggested diagnoses that were not definitive).

## Datasets

We collected all 7102 CPCs published between October 25, 1923 and February 27, 2025. Because CPCs published before 1945 have not been digitized, we parsed these cases from PDF to structured text using a vision–language model (Fig. 3; see Supplementary Information for details). We also collected all 1021 *NEJM* Image Challenge cases published between October 13, 2005 and May 22, 2025, used in a prior study[26].

## Physician Annotation

Ten physicians (five board-certified attending internal medicine physicians, five internal medicine residents) annotated 20 diagnostic case reports each, covering all cases from January 2023 through February 2025 plus a uniform sample from 1991 onward, using a custom web platform. Each case was double annotated to enable pairwise inter-rater comparison. The annotations capture distinct clinical events, or touchpoints—contiguous segments of the vignette that capture the patient at a specific time and place undergoing evaluation or care (e.g., patient at home 6 months prior). Physicians also recorded word-level event type (e.g., disease events, investigations, management), with investigations further annotated by test result (abnormal, explicitly not abnormal). Physician annotations were used to develop and validate an LLM approach (based on o3-mini) which we used to automatically generate clinical and diagnostic annotations across all cases. Annotation quality and similarity between clinicians and LLMs were assessed using multiple evaluation metrics (details in Supplementary Information).

## CPC-Bench Creation and Validation

CPC-Bench consists of the following ten tasks:

1. **Differential Diagnosis (DDx):** DDx from initial presentation, text-only.

2. **Testing Plan:** Testing and management plan from initial presentation.
3. **Literature Search:** Find the accurate citation for a clinical claim.
4. **Diagnostic Touchpoints:** DDx for key events in the patient course, from the initial presentation. Key event boundaries are selected using the physician-validated case annotations.
5. **Question-Answering (QA):** Multiple-choice questions about a patient's initial presentation.
6. **Clinical Reasoning:** Confirmatory/disconfirmatory evidence for each diagnosis on the DDx.
7. **Information Omission:** DDx based on the initial presentation, with normal findings omitted.
8. **Visual Question-Answering (VQA):** Multiple-choice clinical image interpretation, derived from CPC captions.
9. **Visual Differential Diagnosis:** DDx from initial presentation, images and tables only.
10. ***NEJM* Image Challenge:** Multiple-choice question about a clinical image and vignette.

For all tasks except QA, VQA, the *NEJM* Image Challenge, and Literature Search, we used all 409 CPCs published between January 2015 and February 27, 2025. From these, we selected all 377 cases that included a “Differential Diagnosis” section. For the Visual Differential Diagnosis task, we further subset to 356 cases that had at least one figure or table. The VQA dataset was constructed from all 19,968 image-caption pairs from all CPCs published from the year 1980 onwards. The QA and Literature Search tasks are derived from all 937 diagnostic cases published from the year 2000 onwards. Complete details on allowed inputs/outputs, evaluation metrics, and evaluation splits for the public and private datasets are shown in Table 2. See “Benchmark Creation Prompts” in the Supplementary Information for details on the creation of each task. Since some of the underlying cases may be in the training data, we also analyze the performance before and after training cutoffs for all models (Supplementary Table 2).

For each benchmark, we enforced structured outputs from LLMs using task prompts, available in the Supplementary Information. The exception is for CaBot on the Differential Diagnosis task; CaBot was prompted to provide a complete free-text differential diagnosis with citations, using the identical setup from our physician evaluation. We then used an LLM to extract up to 10 diagnoses from the unstructured text for scoring.

We used an LLM judge (GPT-4.1) to assess whether the correct diagnosis appeared in (1) Differential Diagnosis, (2) Diagnostic Touchpoints, (3) Information Omission, and (4) Visual Differential Diagnosis. The final diagnosis was defined as the last section containing the term "diagnosis." To validate the LLM judge, we compared its scoring on 1,467 physician-annotated differentials (3 physicians, prior studies) and found high concordance (accuracy 86%, F1 89%; Supplementary Fig. 5A). For the (2) Testing Plan task, GPT-4.1 evaluated alignment between predicted and actual tests (prompts in Supplementary Information), using 112 physician scores (2 physicians, prior study). Again, concordance was high (accuracy 84%, weighted F1 83%; Supplementary Fig. 5B).

The Literature Search benchmark was generated by extracting all inline citations and corresponding references from each CPC, using an LLM for further quality filtering (see Supplement for prompt), followed by deduplication by title using fuzzy matching and a similarity threshold of 90%[27]. Benchmark questions for the Question-Answering (QA) and Visual Question-Answering (VQA) task are LLM-generated using text from the case (o4-mini, see Supplementary Information). Two physicians (A.R. and J.K.) independently assessed the quality of 50 random QA and VQA examples. In the QA set, both reviewers judged 1 item (2%) to be invalid. In the VQA set, 1 item (2%) was deemed invalid because it had two correct answers, and 2 items (4%) were considered trivial (e.g., requiring counting number of arrows in an image).

We also benchmarked image interpretation tasks across clinical specialties, such as dermatology, radiology, cardiology, and histology. For the *NEJM* Image Challenge cases, we used annotations by a board-certified dermatologist for 764 cases.[28] For the LLM-generated

VQA dataset, we prompted a multimodal LLM to classify the image and caption as one of ten clinical categories (Supplementary Information).

**Statistical Analysis**

95% confidence intervals for LLM performance were computed using the Clopper-Pearson method. Confidence intervals for absolute change in performance over clinical touchpoints were computed using a bootstrap with 1,000 replicates. 95% confidence intervals for pooled performance for multimodal and image-based tasks were computed using a Wald interval.

Fig. 1: Creating Dr. CaBot, the AI Discussant

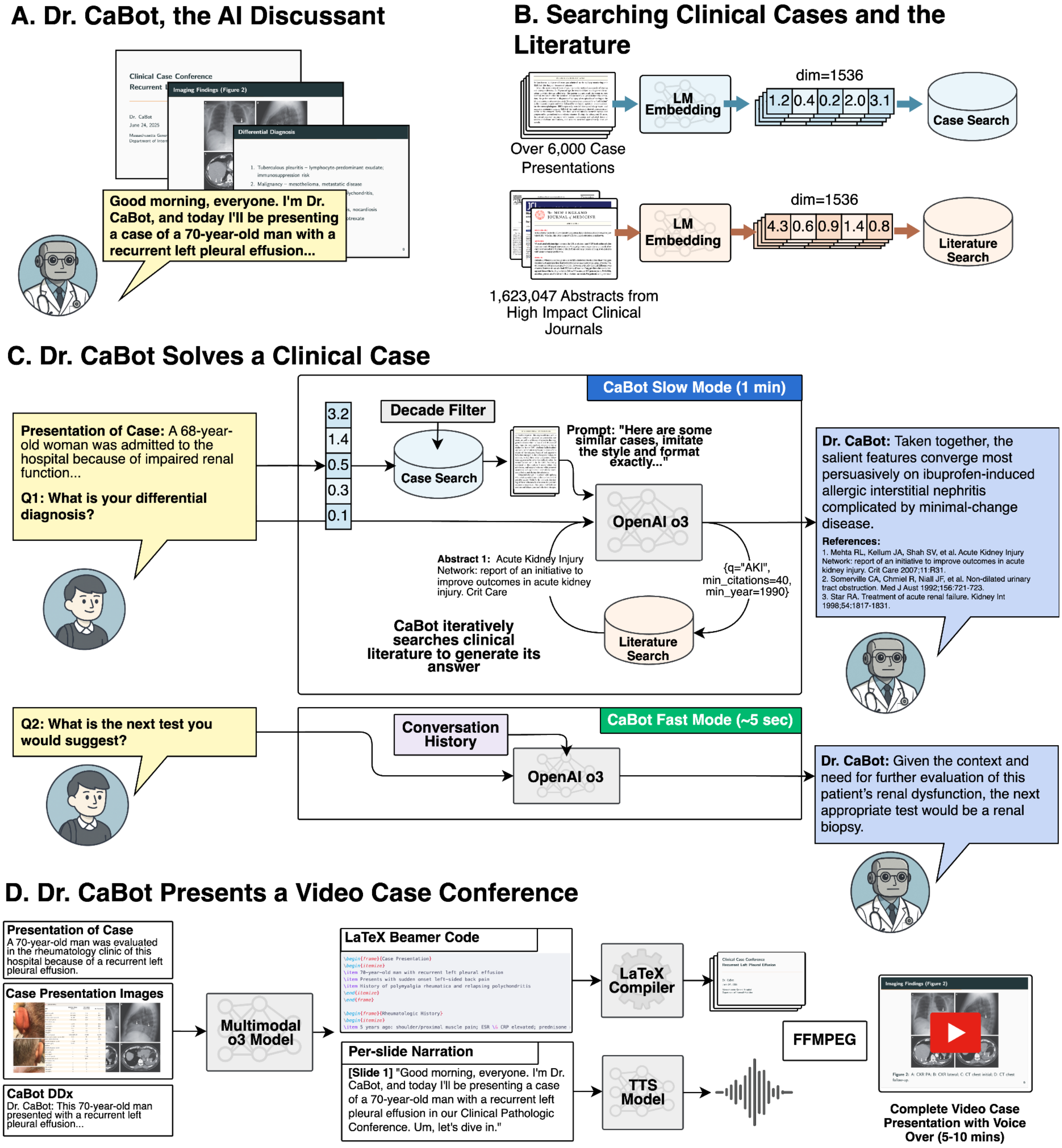

**Fig. 1: Creating Dr. CaBot, the AI Discussant**
**A.** Dr. CaBot is a multimodal AI discussant that can generate video or text-based differential diagnosis in the style of an expert physician discussant. **B.** We create embeddings of the case presentation of over 6,000 CPCs and ~1.6 million clinical abstracts using an OpenAI text embedding model. **C.** At inference time, we embed the input case and provide the model two similar cases (using embedding similarity) from a user-selected decade. The model is prompted to mimic the style of the corresponding differential diagnosis section. The model then iteratively searches the literature to provide a well-sourced differential diagnosis. Follow-up questions can be answered in fast mode without iterative search. **D.** CaBot can generate video presentations in the style of an expert CPC discussant, including narration with slides, after being provided only the presentation of case section.

# Fig. 2: Blinded Physician Evaluations of CaBot Performance

## A. Qualitative Examples of CaBot vs Expert Discussant

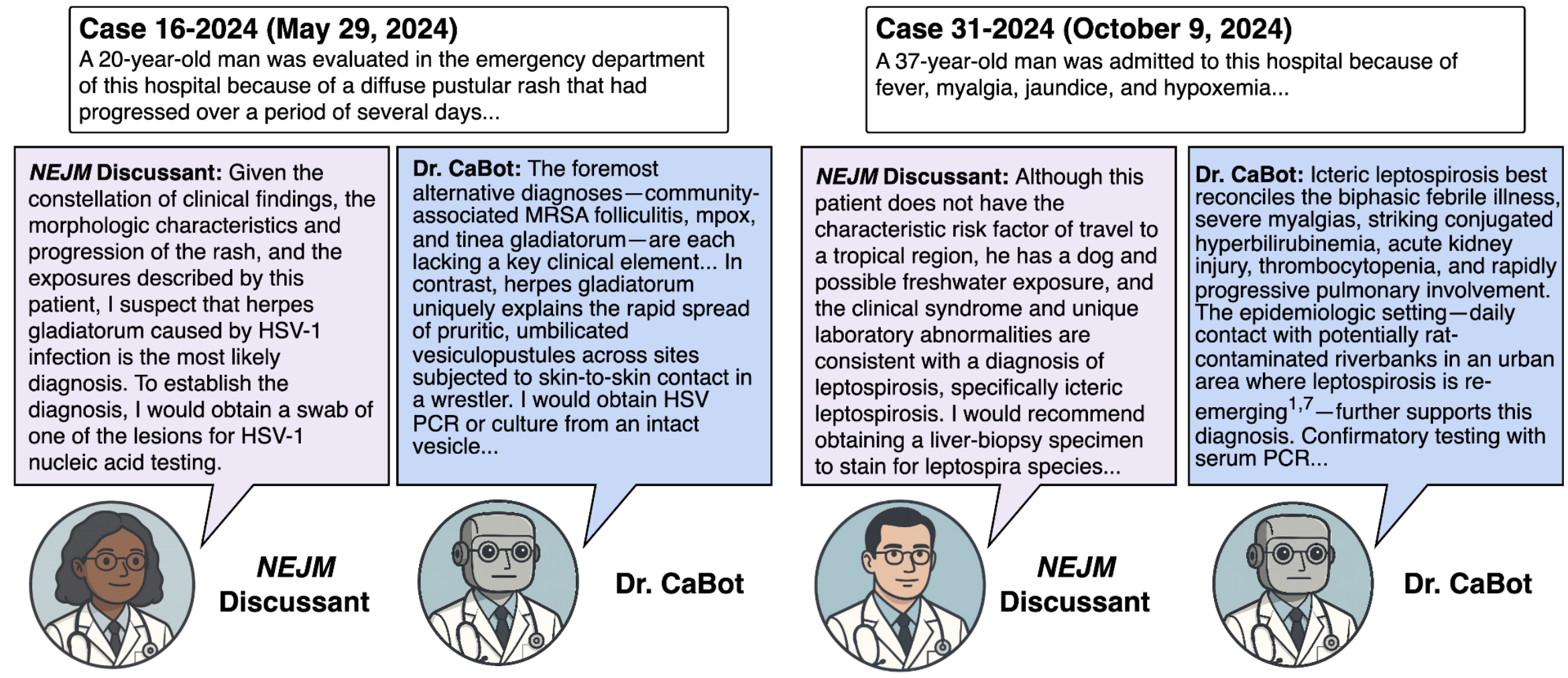

## B. Physician Accuracy in Classifying AI-generated vs Physician-generated Discussion

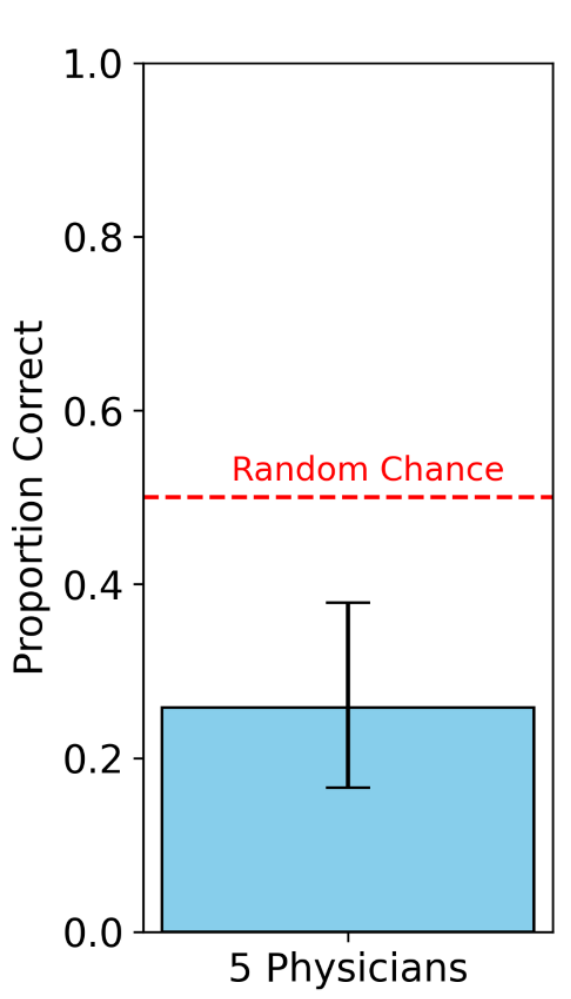

## C. Physician Quality Ratings

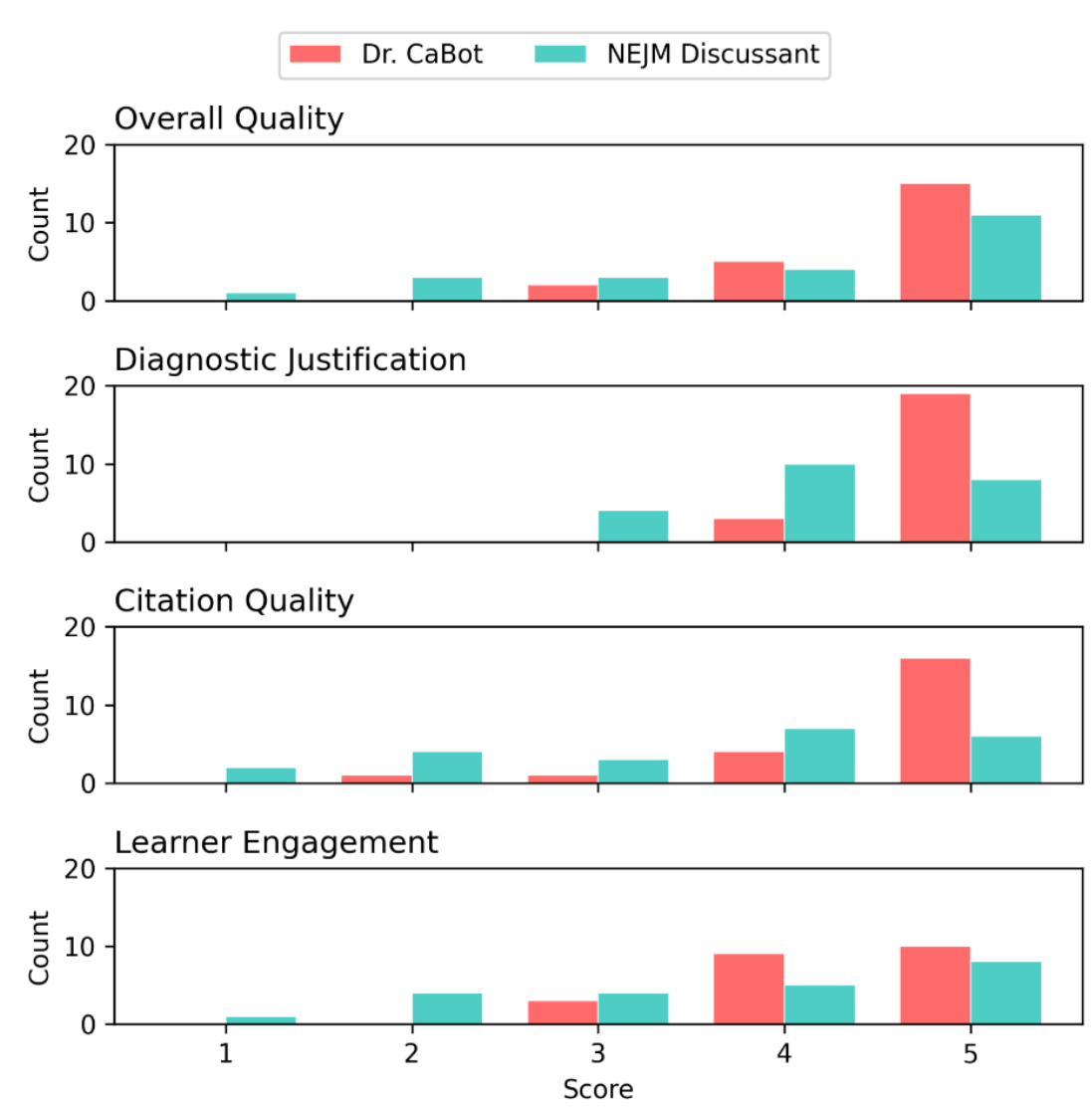

**Fig. 2: Blinded Physician Evaluations of CaBot Performance**

**A.** Examples comparing the differential diagnosis published in the *New England Journal of Medicine* to those generated by the AI discussant, Dr. CaBot. **B.** Mean accuracy of 5 internal medicine physicians in identifying the source of a differential diagnosis as either AI or physician generated. 95% Confidence intervals generated using Wilson method.

**C.** Distribution of quality ratings for the differential diagnosis published in the CPC compared to those generated by Dr. CaBot. Scores were provided for 44 total differential diagnoses by five internal medicine physicians who were blinded to the source of the differential.

# Fig. 3: CPC-Bench: Evaluating LLMs on Text and Multimodal Reasoning Tasks

## A. CPC-Bench Overview

### Dataset Statistics

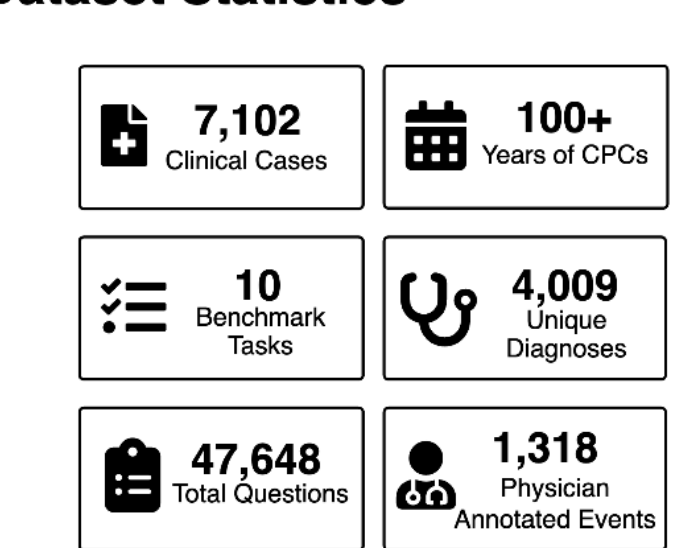

### Clinicopathologic Case (CPC) Format

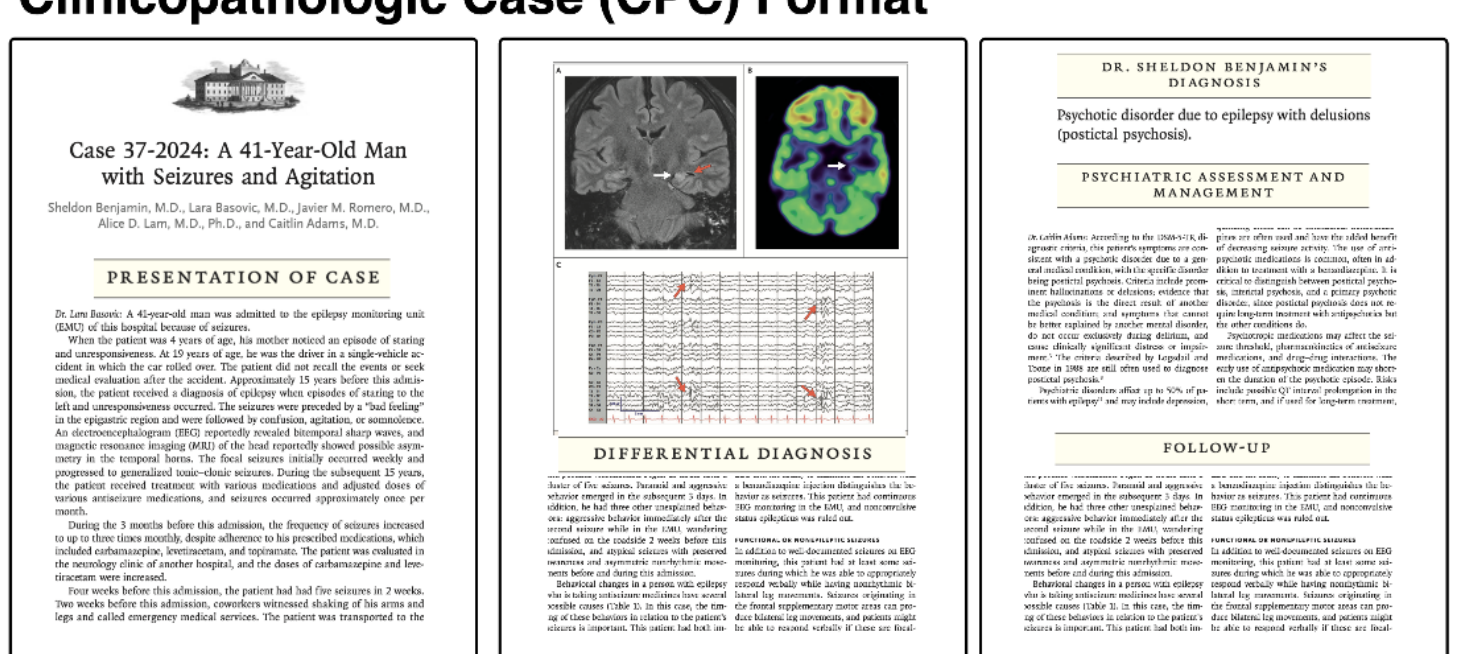

## B. Creation of CPC-Bench

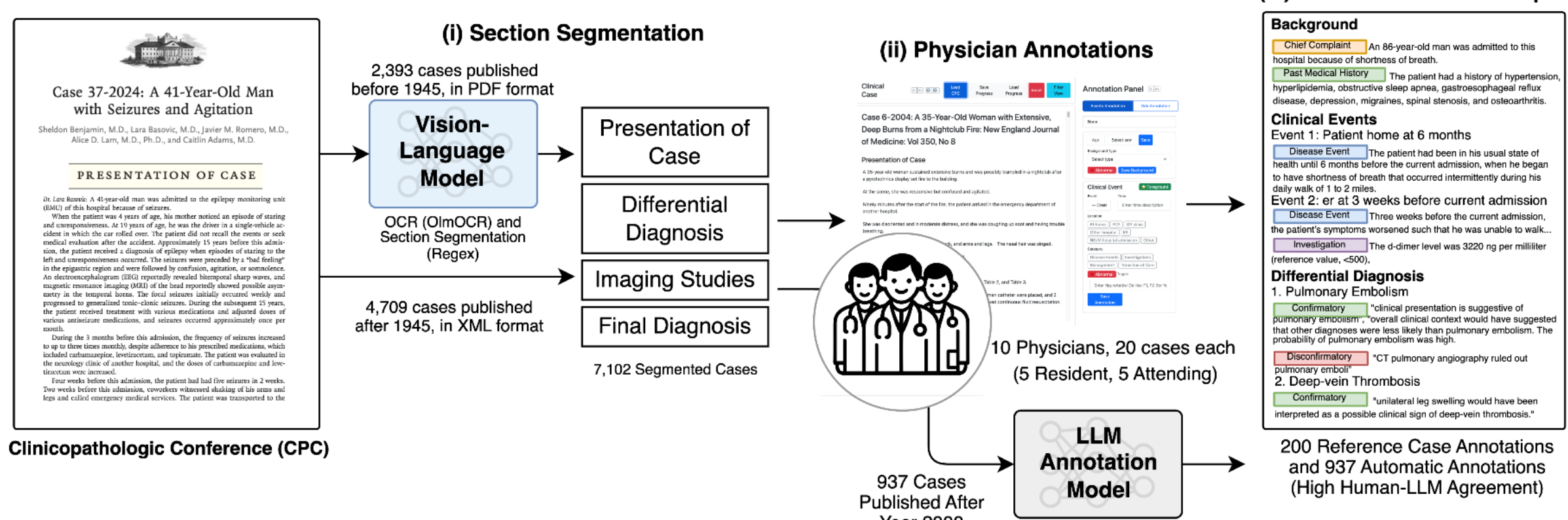

## C. Example Tasks in CPC-Bench

### Text-based Clinical Challenges

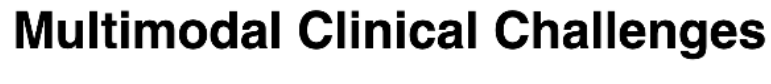

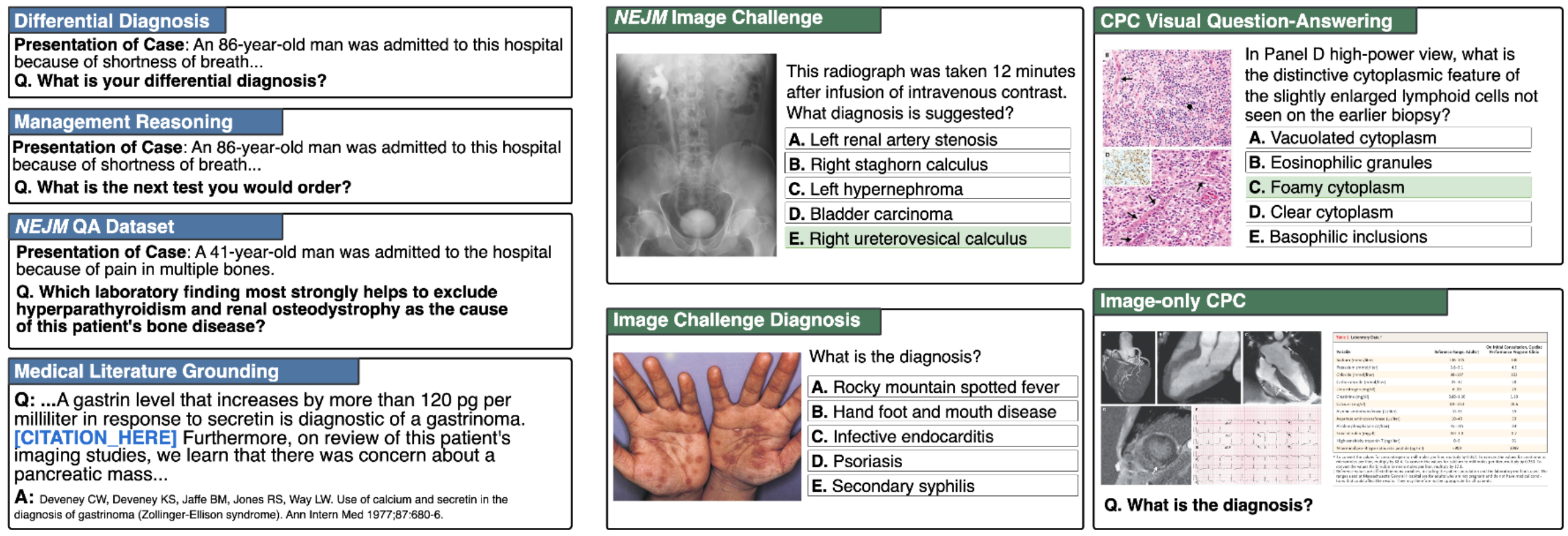

**Fig. 3: CPC-Bench: Evaluating LLMs on Text and Multimodal Reasoning Tasks. A.** Overview of CPC-Bench and the *NEJM* Clinicopathological Conferences (CPCs), which is a series of educational case reports documenting complex cases seen at the Massachusetts General Hospital (MGH). Cases begin with two sections: an initial case presentation (Presentation of Case) and a differential diagnosis provided by an expert physician discussant (Differential Diagnosis). The case then documents follow-up testing, imaging studies, culminating in a final diagnosis. **B.** We developed a custom pipeline to extract structured content from each CPC, had expert physicians annotate cases for diagnostic touchpoints, and developed multiple downstream tasks. **C.** Examples of text-based and multimodal tasks in CPC-Bench.

## Fig. 4: Performance of CaBot and Frontier Models on CPC-Bench

### A. Performance of LLMs on Text-based Tasks

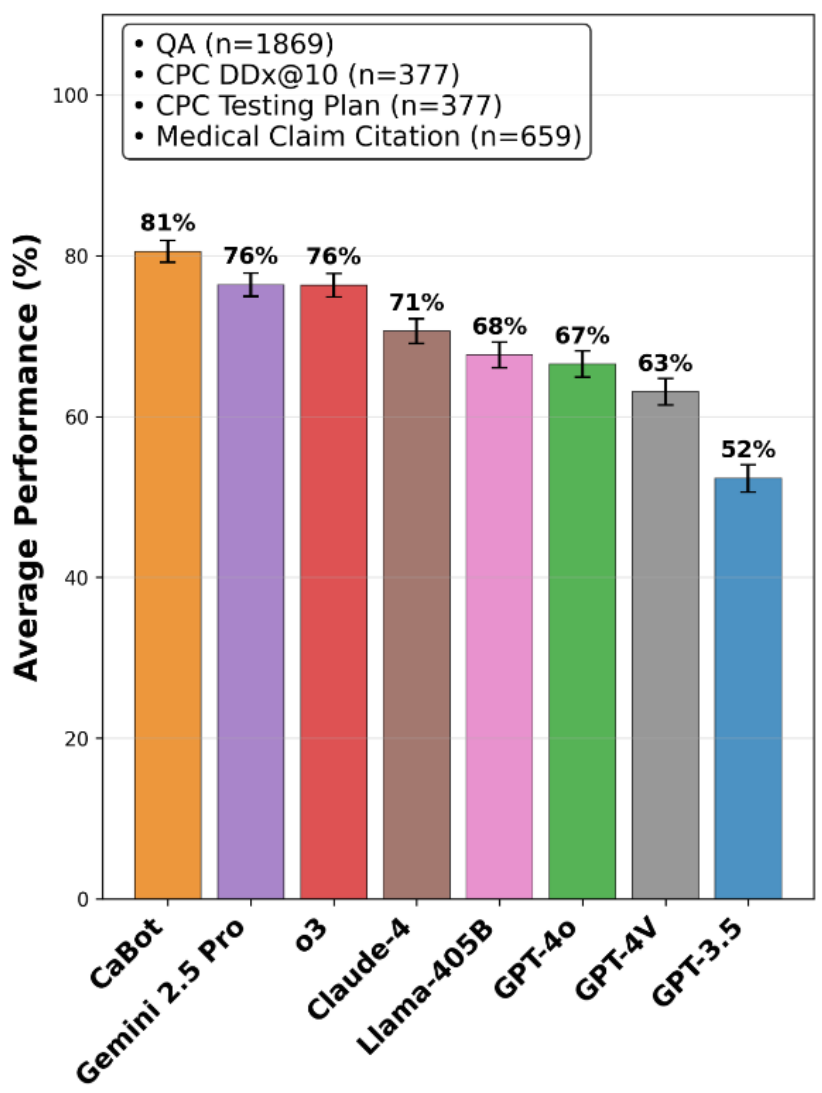

### B. Performance of LLMs on Multimodal Tasks

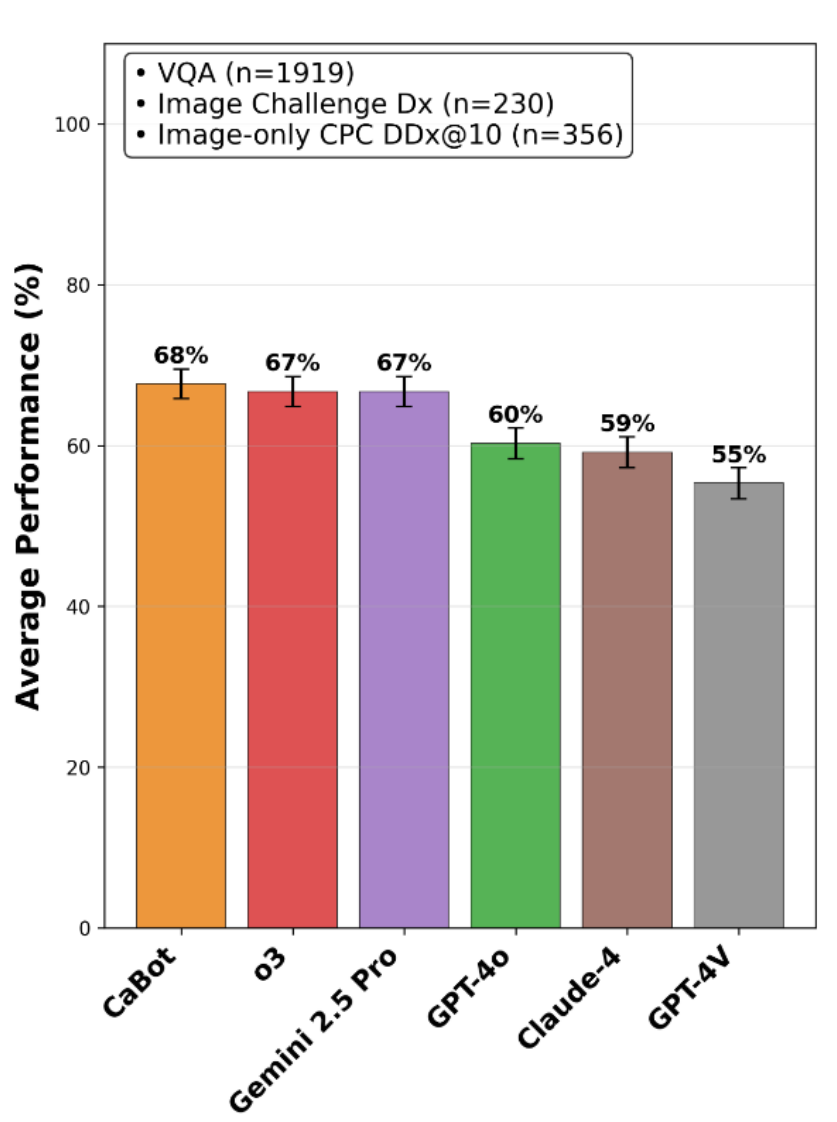

### C. LLM Diagnosis at Key Clinical Touchpoints

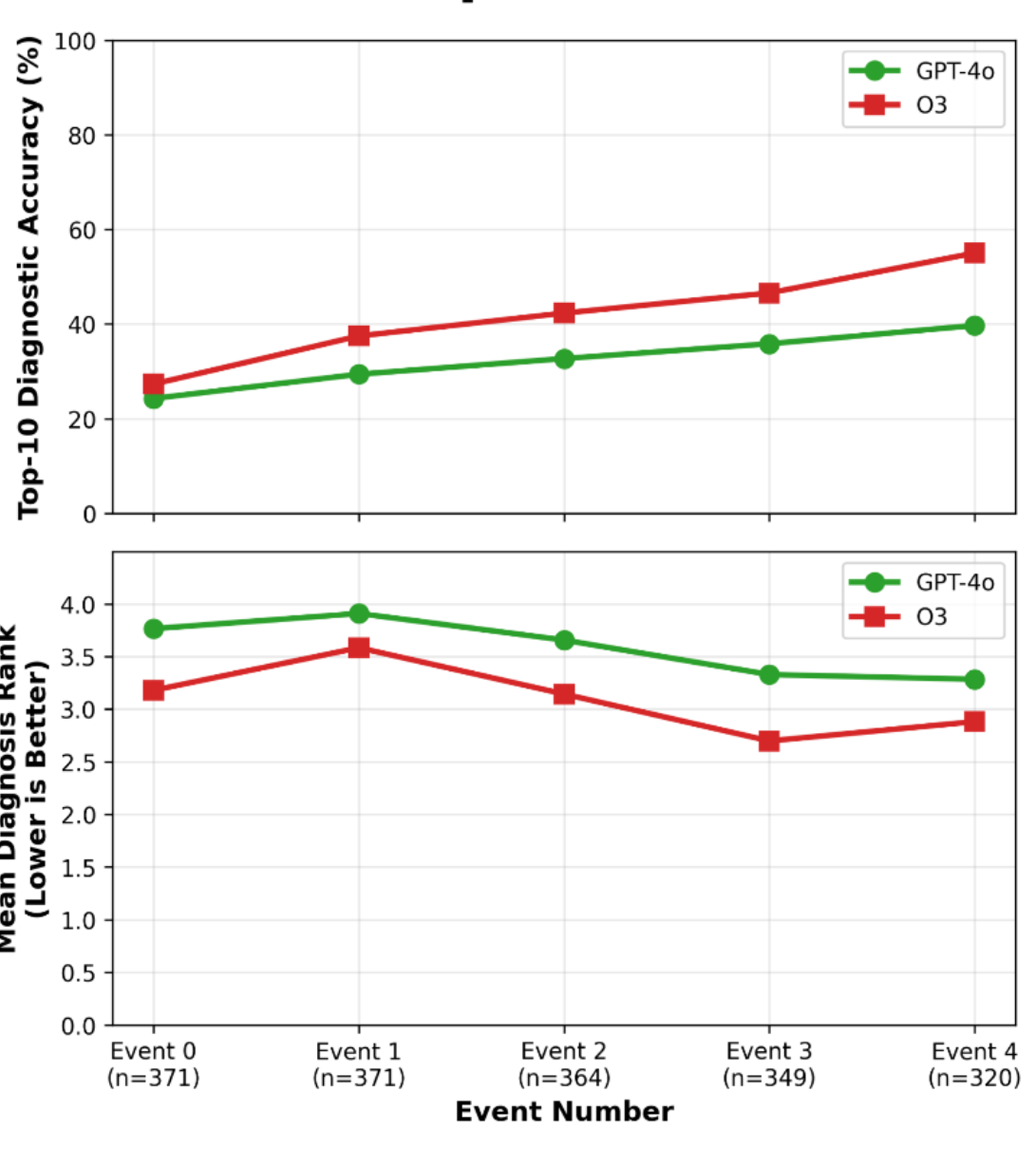

### D. Clinical Literature Search

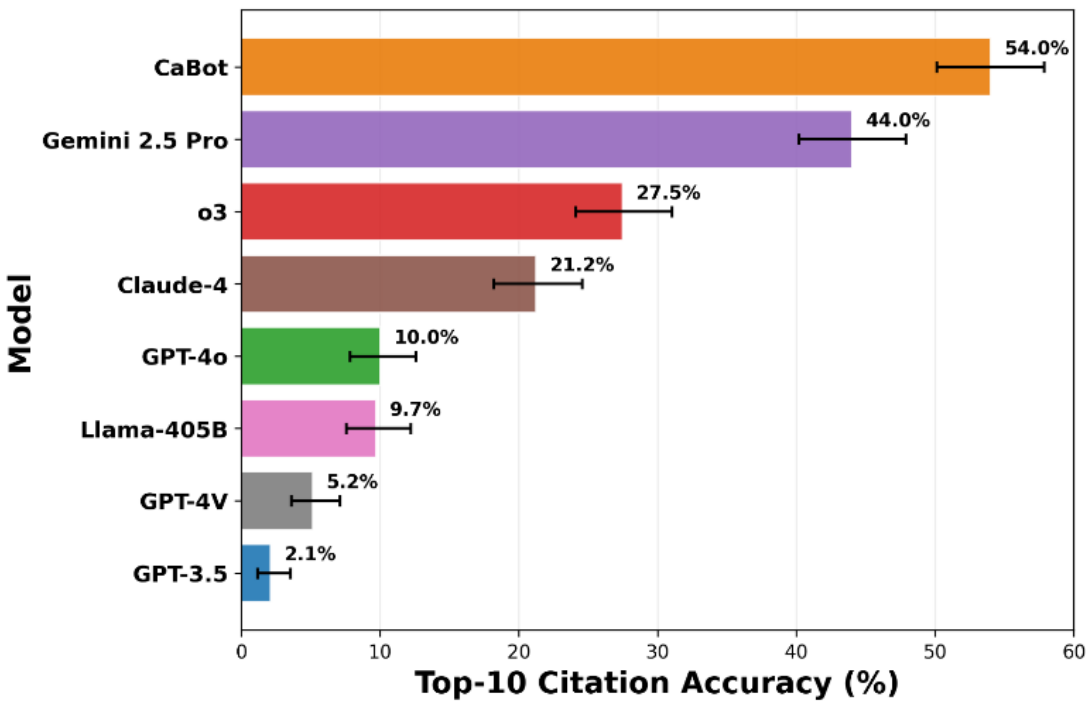

### E. Multimodal Performance by Medical Imaging Specialty

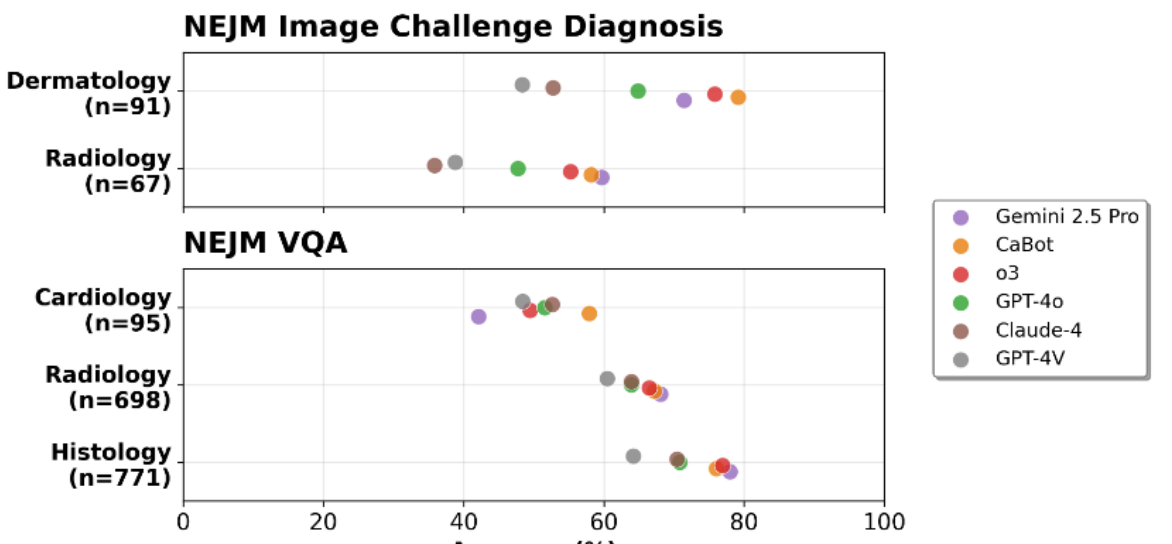

**Fig. 4: Performance of CaBot and Frontier Models on CPC-Bench.**
**A.** Overall performance of CaBot and frontier LLMs on text-based tasks in CPC-Bench. We estimated aggregate performance for each model by taking the micro-average across tasks. 95% confidence intervals are computed using a normal (Wald) approximation. **B.** Performance of CaBot and frontier LLMs on multimodal tasks. **C.** We split the initial case presentation into key clinical events in the patient course (ex. "Home during 3 months before admission") and measured performance of GPT-4o and o3 in suggesting a differential diagnosis at each of these events, based on the final diagnosis. **D.** LLM accuracy in retrieving the correct citation for a medical claim. 95% confidence intervals computed using Clopper-Pearson. **E.** Performance of LLMs on medical imaging by specialty.

## Table 1: Evaluating CaBot on Previously Undiagnosed Rare Disease Cases

| | All (n=72) | Diagnosed by UDN clinical/research evaluation (n=26) | Suggested diagnosis, not fully evaluated (n=46) |
|---|---|---|---|
| Correct disease category by CaBot | 50/72 (69.4%) | 16/26 (61.5%) | 34/46 (73.9%) |
| Specific diagnosis or gene named by CaBot | 46/72 (63.9%) | 13/26 (50.0%) | 33/46 (71.7%) |

**Table 1: Evaluating CaBot on Previously Undiagnosed Rare Disease Cases.** Summary of reviewer assessments for the 72 Undiagnosed Diseases Network (UDN) cases retained for analysis. Cases are stratified by whether the UDN evaluated the patient and identified a diagnosis (n = 26) or the UDN suggested a diagnosis but the patient was not fully evaluated (n = 46). Each CaBot letter was scored against the confirmed or suggested diagnosis. Each CaBot response was reviewed by certified genetic counselors (K.L., E.G.).

# Table 2: Overview of CPC-Bench Tasks

| Name | Task | Inputs | Output | Evaluation Method | CPC-Bench Leaderboard (Private) | CPC-Bench Public (to be released) |
|---|---|---|---|---|---|---|
| 1. Differential Diagnosis (DDx) | DDx from initial presentation, text-only | - All text from the “Presentation of Case” section | Ranked list of 10 diagnoses. Each diagnosis is listed as free-text. | Physician-validated LLM judge is used to determine the presence and rank of the final diagnosis. | All 377 Diagnostic CPCs, January 2015 to February 27, 2025 | 100 Diagnostic CPCs, 2000 to 2025. |
| 2. Testing Plan | Testing and management plan from initial presentation | - All text from the “Presentation of Case” section | Free-text description of the next test or management plan. | Physician-validated LLM judge is used to score the following rubric:<br>2: The testing plan suggested by the student matches, or is a superset of, the testing plan actually performed.<br>1: The testing plan suggested by the student is different, but would have been helpful in finding the final diagnosis.<br>0: The tests suggested by the student would not have been helpful at all in finding the final diagnosis. | All 377 Diagnostic CPCs, January 1st, 2015 to February 27, 2025 | 100 Diagnostic CPCs, 2000 to 2025. |
| 3. Literature Search | Find a relevant citation for a clinical claim | - Snippet of 500 characters before and after a masked citation reference, from an *NEJM* CPC.<br><br>- Year in which the claim was made | Ranked list of 10 citations, each containing the following fields:<br>- Rank<br>- Title<br>- First Author<br>- Publication Year<br>- Journal | An LLM judge is used to check for a complete match between the correct reference and each of the predicted references. | 937 diagnostic cases published from January 1st, 2000 to February 27, 2025. 5,286 total examples. Evaluated on a 20% held-out test set, split by case (n=659) | 100 Diagnostic CPCs, 2000 to 2025. |
| 4. Diagnostic Touchpoints | DDx for key events in the patient course, from the initial presentation | - Incremental text snippets from the Presentation of Case, with at most 5 rounds. | Ranked list of 10 diagnoses. Each diagnosis is listed as free-text. | Physician-validated LLM judge is used to determine the presence and rank of the final diagnosis. | All 377 Diagnostic CPCs, January 2015 to February 27, 2025 | 100 Diagnostic CPCs, 2000 to 2025. |
| 5. Question-Answering (QA) | Multiple-choice questions about a patient’s initial presentation | - Question<br>- 5 multiple-choice options | Letter of the correct answer choice | Exact-match | 937 diagnostic cases published from January 1st, 2000 to February 27, 2025. 18,600 total examples. Evaluated on a 10% held-out test set, split by case (n=1,869) | 100 Diagnostic CPCs, 2000 to 2025. |
| 6. Clinical Reasoning | Confirmatory/disconfirmatory evidence for each diagnosis on the DDx | - All text from the “Presentation of Case” section | Ranked list of 10 diagnoses, each containing a list of confirmatory and disconfirmatory evidence. Evidence is provided in free-text. | We measure the precision and recall:<br>$\text{Recall} = \frac{\lvert R_{\text{match}} \rvert}{\lvert R_{\text{gt}} \rvert} \text{Precision} = \frac{\lvert R_{\text{match}} \rvert}{\lvert R_{\text{pred}} \rvert}$<br>- $R_{gt}$: The set of triples (diagnosis, evidence, confirm/disconfirm) extracted from the “Differential Diagnosis” section<br>- $R_{pred}$: The set of triples (diagnosis, evidence, confirm/disconfirm) generated by the LLM.<br>- $R_{match}$: The set of triples where the LLM and the expert agree on the diagnosis, the evidence, and the confirm/disconfirm relation, as judged by another LLM. | All 377 Diagnostic CPCs, January 2015 to February 27, 2025 | 100 Diagnostic CPCs, 2000 to 2025. |

| 7. Information Omission | DDx based on the initial presentation, with normal findings omitted | - Text from the “Presentation of Case” section, excluding text labeled as “normal” findings | Ranked list of 10 diagnoses in free-text. | Physician-validated LLM judge is used to determine the presence and rank of the final diagnosis. | All 377 Diagnostic CPCs, January 2015 to February 27, 2025 | 100 Diagnostic CPCs, 2000 to 2025. |
|---|---|---|---|---|---|---|
| 8. Visual Question-Answering (VQA) | Multiple-choice clinical image interpretation, derived from *NEJM* CPC captions | - Question<br>- 5 multiple-choice options | Letter of the correct answer choice | Exact-match | All 1984 diagnostic CPCs published between January 1st, 1980 and February 27, 2025. 19,786 total examples. Evaluated on a 10% held-out test set, split by case (n=1,919) | 100 Diagnostic CPCs, 2000 to 2025. |
| 9. Visual Differential Diagnosis | DDx from initial presentation, images and tables only | - Images and tables from the initial presentation | Ranked list of 10 diagnoses. Each diagnosis is listed as free-text. | Physician-validated LLM judge is used to determine the presence and rank of the final diagnosis. | All 356 diagnostic cases published between January 2015 and February 27, 2025 that have at least one figure or table. | 100 Diagnostic CPCs, 2000 to 2025. |
| 10. *NEJM* Image Challenge | Multiple-choice question about a clinical image and vignette | - Image<br>- Question<br>- 5 multiple-choice options | Letter of the correct answer choice | Exact-match | All 1021 *NEJM* Image Challenge cases, published between October 13, 2005 and May 22, 2025. | 30 Image Challenge Questions |
| 10a. *NEJM* Image Challenge (Diagnosis) | Subset where no vignette is provided | - Image<br>- Question<br>- 5 multiple-choice options | Letter of the correct answer choice | Exact-match | 230 *NEJM* Image Challenge cases. | 100 Diagnostic CPCs, 2000 to 2025. |

**Table 2: Overview of Tasks.** Validation details of the LLM judge, and the complete prompts and JSON schema for each benchmark are available in the Supplementary Information.

## Acknowledgments

We gratefully acknowledge support from the Dunleavy Fund for Clinical AI at Harvard Medical School (TAB), the National Institutes of Health/National Institute of Environmental Health Sciences R01ES032470 (AKM), the National Institute of Neurological Disorders and Stroke of the National Institutes of Health through Award U2CNS132415 (KL, EG), and the Harvard Medical School Dean's Innovation Awards for the Use of Artificial Intelligence in Education, Research, and Administration (AKM). We thank NEJM Group and the Massachusetts Medical Society for permission to use published content in this study. The content is solely the responsibility of the authors and does not necessarily represent the official views of the National Institutes of Health.